\documentclass[twoside,british,a4paper,twocolumn]{article}
\usepackage{amsmath}
\usepackage{amsfonts}
\usepackage{amssymb,upref}
\usepackage{tgheros} %
\usepackage{tgtermes} %
\usepackage{anyfontsize}
\usepackage{xcolor}
\usepackage{graphicx}
\usepackage{enumitem}
\usepackage{subfig}
\usepackage{siunitx}
\usepackage{multirow}

\usepackage{balance}

\usepackage[utf8]{inputenc}
\usepackage[T1]{fontenc}

\usepackage{isodate}
\cleanlookdateon

\usepackage{lettrine}
\pdfmapfile{=montserrat.map}

\usepackage{lipsum}

\usepackage{newtxmath}

\usepackage{geometry}
\usepackage{fancyhdr}

\fancypagestyle{firstpagestyle}
\usepackage{titlesec}
\titleformat{\section}[block]{\bfseries\upshape\Large\sffamily\boldmath}{}{0.em}{}
\titlespacing*{\section}{0pt}{0.8em plus 0ex minus 0ex}{0em plus 0.ex}

\usepackage{titling}
\pretitle{\noindent\begin{minipage}{\textwidth} \raggedright\huge\bfseries\upshape\sffamily\boldmath
\color[rgb]{0,0,0.8}
}
\posttitle{\end{minipage} \vskip 0.6em}
\preauthor{\noindent \begin{minipage}{\textwidth} \large\mdseries\sffamily}
\postauthor{
  \end{minipage} 
  \vskip 0.4em
  {
   \sffamily
   \color{black!80}
   \noindent \small
   \address
   \par %
   \authoremail
  }
   \par \vskip 0.4em
}
\predate{}
\postdate{}
\date{}

\usepackage{caption}
\DeclareCaptionFont{cfs}{\fontsize{8.5}{10.25}\selectfont}
\DeclareCaptionLabelSeparator{vline}{\;|\;}
\usepackage{tcolorbox}
\definecolor{abstractboxcolor}{cmyk}{0.1,0,0,0}
\newtcolorbox{abstractbox}{
  arc=0pt,
  boxrule=0pt,
  colback=abstractboxcolor,
  boxsep=0.5em,
  left=0pt, right=0pt, bottom=0pt, top=0pt,
  width=\columnwidth
}
\makeatletter
 \def\@textbottom{\vskip \z@ \@plus 1pt}
 \let\@texttop\relax
\makeatother
\renewenvironment{abstract}{
   \noindent
   \begin{minipage}{\textwidth}
   \upshape\sffamily \bfseries
   \fontsize{9}{11.5}\selectfont
  }{
   \end{minipage} 
   \vskip 2.0em
  }

\usepackage[super,numbers,sort&compress]{natbib}

\makeatletter \def\NAT@def@citea{\def\@citea{\NAT@separator\,}} \makeatother %
\usepackage{etoolbox}
\apptocmd{\sloppy}{\hbadness 10000\relax}{}{}

\usepackage[%
  bookmarks=true,
  colorlinks,
  linkcolor=blue,
  urlcolor=blue,
  citecolor=blue,
  plainpages=false,
  pdfpagelabels,
  final,
  breaklinks=true
]{hyperref}
\hypersetup{
pdftitle={Improved Drug-target Interaction Prediction with Intermolecular Graph Transformer}, 
pdfauthor={},
pdfkeywords={}
}

\usepackage{booktabs}
\newcommand{\angstrom}{\textup{\AA}}
\graphicspath{ {./figures/}  }

\title{Improved Drug-target Interaction Prediction with Intermolecular Graph Transformer}
\newcommand\shorttitle{IGT}

\author{
Siyuan Liu\textsuperscript{1,2,3,\#},
Yusong Wang\textsuperscript{4,3,\#},
Tong Wang\textsuperscript{3,*},
Yifan Deng\textsuperscript{3,5},
Liang He\textsuperscript{3},
Bin Shao\textsuperscript{3},
Jian Yin\textsuperscript{1,2},
Nanning Zheng\textsuperscript{4},
Tie-Yan Liu\textsuperscript{3}
}
\newcommand\shortauthor{Liu et al.}

\newcommand\address{\textsuperscript{1}School of Computer Science and Engineering, Sun Yat-sen University, Guangzhou, 510006, China

\textsuperscript{2}Guangdong Key Laboratory of Big Data Analysis and Processing, Guangzhou, 510006, China

\textsuperscript{3}Microsoft Research Asia, Beijing, 100080, China

\textsuperscript{4}Institute of Artificial Intelligence and Robotics, Xi’an Jiaotong University, Xi’an, 710049, China

\textsuperscript{5}School of Computer Science, Fudan University, Shanghai, 200433, China

\textsuperscript{\#}Equal contribution}

\newcommand\authoremail{$^*$watong@microsoft.com}

\begin{document}

\twocolumn[{
\begin{@twocolumnfalse}

\maketitle
\thispagestyle{firstpagestyle}

\begin{abstract}
  The identification of active binding drugs for target proteins
  (termed as drug-target interaction prediction)
  is the key challenge in virtual screening, which plays an essential role in drug discovery.
  Although recent deep learning-based approaches achieved better performance than
  molecular docking, existing models often neglect certain aspects %
  of the intermolecular information, hindering the performance of prediction.
  We recognize this problem and propose a novel approach named Intermolecular Graph Transformer (IGT)
  that employs a dedicated attention mechanism to model intermolecular information with a three-way
  Transformer-based architecture.
  IGT outperforms state-of-the-art approaches by
  9.1\% and 20.5\% over the second best for binding activity and
  binding pose prediction respectively, and shows superior
  generalization ability to unseen receptor proteins.  Furthermore,
  IGT exhibits promising drug screening ability against SARS-CoV-2 by
  identifying 83.1\% active drugs that have been validated by wet-lab
  experiments with near-native predicted binding poses.%
\end{abstract}

\vspace{-2mm}

\end{@twocolumnfalse}
}]

\section{Introduction}

The war between diseases and human beings has never ended. When a
new disease such as COVID-19 breaks out, \emph{de novo} drug design is not always the best option
due to the huge cost in both expense and time~\citep{da2019virtual,lin2020review,wu2020analysis}.
There are patients suffering every minute when no effective
drug is available, while using unknown drugs might cause unpredictable
consequences to the patients.  In such circumstances, screening drugs from
known chemical compounds, followed by cell experiments and clinical
trials, is a better
alternative~\citep{ashburn2004drug,novac2013challenges,xue2018review,dudley2011exploiting}.

With the rapid development of computational power,
\emph{in silico} drug screening that predicts the drug-target interactions (DTI)
to identify active binding drug candidates
has become one of the most important techniques in drug discovery~\citep{da2019virtual,lin2020review,csermely2013structure}.
Among the DTI prediction techniques,
molecular/quantum dynamics simulation achieves high accuracy 
by applying well-designed physical force fields~\citep{de2016role,durrant2011molecular,sledz2018protein}.
However, such simulations are prohibitively expensive to be applied in the high-throughput screening of 
compound libraries~\citep{lin2020review,novac2013challenges,sledz2018protein,de2016role}.
Therefore, an alternative approach called molecular docking is widely adopted
to trade accuracy for higher throughput by applying heuristics and
empirical scoring functions~\citep{sledz2018protein,lin2020review}. 
Molecular docking achieves a much higher throughput, but with the cost of a lower
accuracy compared to the molecular/quantum dynamics approaches~\citep{sledz2018protein,lin2020review}.

Recently, many neural network models have been proposed for DTI prediction,
enabling high accuracy DTI predictions at affordable costs.
These models can be roughly divided into two types.
Given a pair of compound and receptor,
the first type of models (termed as the \emph{type-I} approaches)~\citep{lim2019gnndti,ragoza2017protein,wallach2015atomnet}
first dock the ligand in question to the receptor using molecular docking software,
then extract the structure around the binding site from the resulting pose,
and finally feed the structure into the neural network.
For example, AtomNet~\citep{wallach2015atomnet} embeds the binding site as a 3D grid
and feeds the grid into a 3D convolutional neural network (CNN) to predict binding activity.
Following AtomNet, \citet{ragoza2017protein}
also applies 3D CNN to voxelized binding sites to predict the binding activity and binding pose simultaneously. 
Different from the above CNN models, \citet{lim2019gnndti} represents the binding site
as a graph, and uses a graph neural network (GNN) to predict the binding activity and the binding pose.
Specifically, this model adopts a distance-aware attention to handle intermolecular information,
which greatly improved its performance on the DUD-E benchmark dataset~\citep{mysinger2012dude}.

The second type of models (termed as the \emph{type-II} approaches)~\citep{li2020monn,zheng2020drugvqa,torng2019graph},
in contrast with the type-I approaches, first represent and process the receptor and ligand individually,
and then combine the respective representations for binding activity prediction.
For example, \citet{torng2019graph} employs two separate GNNs for representing the receptor pocket and the ligand,
after which a prediction is made from the concatenation of the two representations.
MONN~\citep{li2020monn} uses a CNN to encode full-length protein sequences to represent receptors
and a GNN to represent the ligand, and the inner product of the two representations is used for
the prediction.

These two types of models can be seen as two different yet complementary ways towards modeling the interactions between
receptors and ligands, which sits at the center of the DTI problem.
The type-I approaches leverage the intermolecular information (i.e., coordinates and distances)
from poses generated by molecular docking software.  This information, albeit possibly inaccurate, reflects certain
human knowledge about the physicochemical aspects of the DTI problem.
However, these approaches usually treat the \emph{inter}molecular edges indifferently as \emph{intra}molecular
edges, which in fact neglects the \emph{topology} information (i.e., which edges are intermolecular).
The type-II approaches choose not to use the generated poses, but instead learn the protein-ligand
interactions with dedicated network modules.
In this way, the prediction of binding activity is constrained
to be a function of the learned interaction rather than the individual receptor and ligand, which also
reflects human inductive bias of the problem.
Nonetheless, these models effectively only use the \emph{topology} aspects of the intermolecular interactions.
In summary, in both these two types of models, certain information about the intermolecular interactions
is neglected and left to be implicitly learned by the neural network.

In view of this, we argue that the receptor-ligand interactions can be better modeled using a combined
approach.
We validate this argument by proposing and experimenting with a novel deep learning model called
``Intermolecular Graph Transformer'' (IGT).
IGT takes the same inputs as the type-I approaches, while bears an architectural resemblance
to type-II approaches.
In particular, the IGT has a three-way Transformer-based architecture~\citep{vaswani2017attention,dwivedi2021generalization}.
In each network block of the IGT, three graphs, namely the receptor graph, the ligand graph and the complex graph,
are first individually processed and finally combined through a specially designed mechanism, termed as
``intermolecular attention''.
We show that this three-way design, as well as the intermolecular attention, can greatly improve IGT's
ability to learn the receptor-ligand interactions through an ablation study comparing the IGT with
a one-way graph transformer counterpart.

We apply IGT to both binding activity prediction and binding pose prediction.
For binding activity prediction,
IGT was trained on two datasets, namely DUD-E
and LIT-PCBA~\citep{tran2020lit}, respectively,
and then evaluated on the test datasets. 
IGT surpassed the best of the state-of-the-art
models on almost all metrics, with 9.1\% and 8.7\% improvements on AUPRC respectively for DUD-E and LIT-PCBA.
These performance gains demonstrate that IGT is able to capture the true physical signals instead of simply fitting to the training samples.
A notable finding from our experiments is that, training on unbiased samples makes the models better for generalization.
When evaluated on the independent test dataset MUV~\citep{rohrer2009maximum}, a 5.3\% increase of AUROC is witnessed, reflecting the better
generalization capability of IGT.
For pose prediction, IGT was trained and evaluated on PDBbind~\citep{liu2017pdbbind}, which drastically outperformed molecular docking
and achieved a 20.5\% relative improvement to the second best.
Finally, to test IGT on real-world DTI applications, we apply both the activity prediction model
and the pose prediction model for drug virtual screening against SARS-CoV-2.
When evaluated on Diamond SARS-CoV-2 drug dataset, %
IGT successfully identified the active drugs with near-native binding poses.

Although we proposed and evaluated IGT as a DTI prediction model, it is worth noting that
IGT can serve as a general modeling framework for other research fields regarding two-body interactions,
such as drug-drug interaction, protein-protein interaction, etc.
We sincerely hope that the IGT, our solution to both binding activity prediction and pose prediction,
could accelerate practical drug discovery applications and inspire new ideas in future research of related fields.

\section{Results}

\subsection{Overview of Intermolecular Graph Transformer}

Intermolecular Graph Transformer (IGT) is a novel graph transformer neural network for DTI task
based on the famous Transformer architecture~\citep{vaswani2017attention} and 
its generalization, Graph Transformer~\citep{dwivedi2021generalization}.
The original Transformer is currently the dominant model in 
natural language processing~\citep{vaswani2017attention} that has been shown
transferable to and successful in other important fields, such as computer vision~\citep{dosovitskiy2020image}.
We adopted a variant of the dot-product attention from Graph Transformer
and applied it simultaneously to three graphs, i.e., the ligand graph, the receptor graph, and the graph for the complex structure.
In addition, the graph dot-product attention layers are interleaved with an \textit{intermolecular attention}, 
a dedicated graph-level operator we designed to better exploit the intermolecular edges in the graphs.

\begin{figure*}[htp]
  \centering
  \includegraphics[width = 0.95\textwidth]{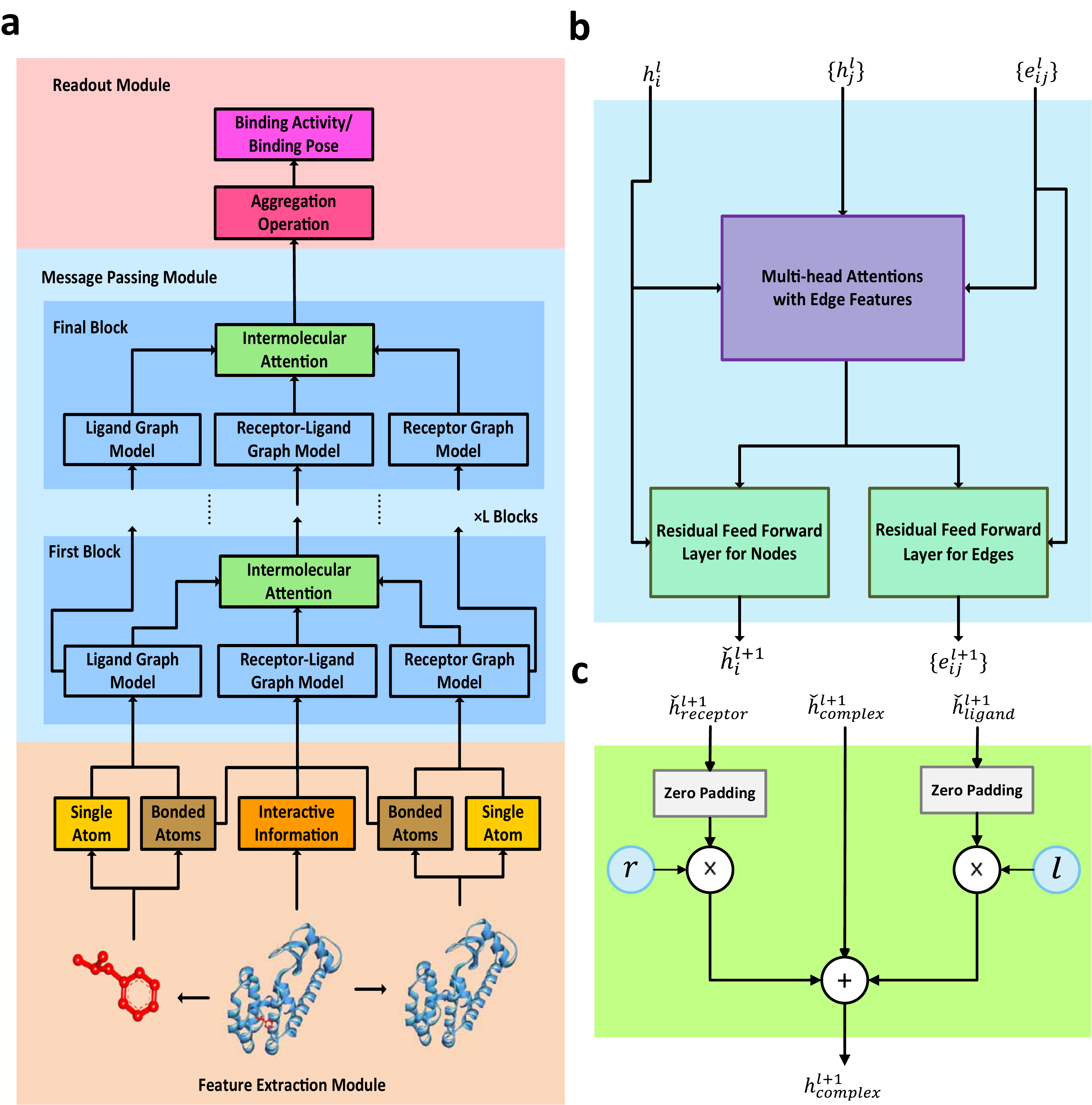}
  \caption{
    The overall model architecture of IGT.
    \textbf{a}. The flowchart of IGT. The feature extraction module,
    the message passing module and the readout module are shown in the orange, blue and pink panels, respectively.
    \textbf{b}. The architecture of graph model in the message passing module. Node features as well as positional encodings
    pass through a multi-head attention module while the edge features act as weights for each
    attention between two nodes. Then node features are updated by residual feed forward layers.
    \textbf{c}. The depiction of intermolecular attention. Attention weights are assigned to the ligand,
    receptor, and complex graphs. The node features of the complex graph are updated by the
    information of all the three graphs.
  }
  \label{fig:igt-arch}
\end{figure*}

In brief, IGT consists of three modules, i.e., 
a feature extraction module, a message passing module, and a readout module (Figure~\ref{fig:igt-arch}a).
The feature extraction module extracts the atom and bond features from the ligand,
the receptor, and the complex structure, respectively. The extracted features are then fed into the
corresponding graphs in the message passing module, which consists of tandem repeated \textit{IGT blocks}.
In each IGT block, we adopt a graph-aware dot-product attention for each graph (Figure~\ref{fig:igt-arch}b) and
an intermolecular attention to aggregate all information to update the complex graph
(Figure~\ref{fig:igt-arch}c).  The node features of three graphs in the final block
are then fed into the readout module.  All messages are aggregated by the aggregation
operation and the score of binding activity or the score of binding pose is predicted.
For more details of IGT, please refer to the Methods section.

\subsection{IGT achieves the best performance on binding activity prediction}
\begin{figure}[htp]
  \centering
  \includegraphics[width = 0.48\textwidth]{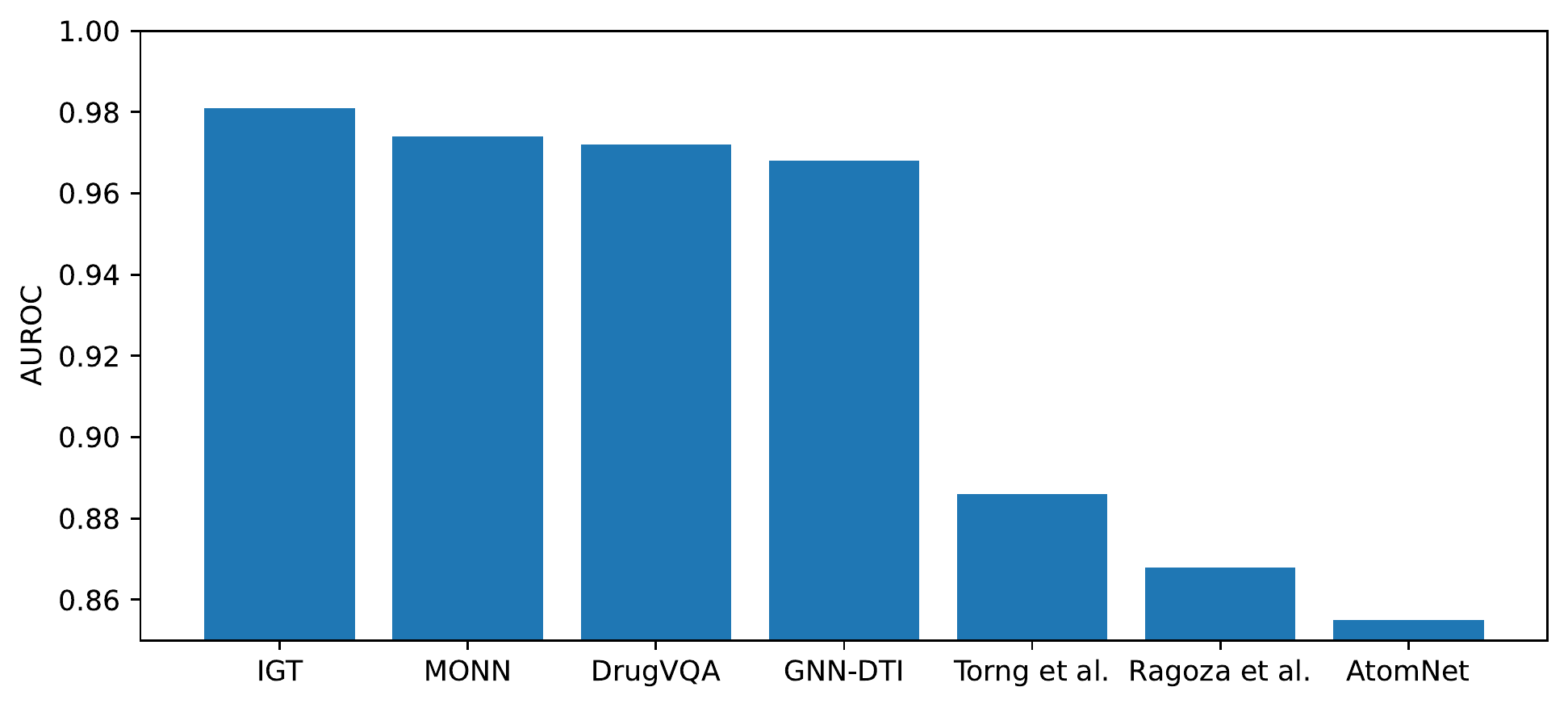}
  \caption{
    Comparison of AUROC on the DUD-E dataset, among IGT and other approaches~\citep{li2020monn,zheng2020drugvqa,lim2019gnndti,torng2019graph,ragoza2017protein,wallach2015atomnet}.
  }
  \label{fig:auroc}
\end{figure}

To evaluate the performance of IGT for binding activity prediction, 
we first used the widely adopted DUD-E dataset as a benchmark. 
We randomly split the DUD-E dataset into training, validation, and test sets by the target proteins with the ratio of 0.70:0.15:0.15.
All compounds were docked by Smina~\citep{koes2013lessons} with the default parameters, which is
an improved docking software based on Autodock Vina~\citep{trott2010vina}.
For binding activity prediction, the docked pose with the lowest energy of each chemical compound was selected for model training.
The IGT was trained on the training and validation sets with a learning rate of 1e-5 (see Methods for more details),
and then evaluated on the test set. As shown in Figure~\ref{fig:auroc}, IGT achieved an remarkable
AUROC of 0.981 on the DUD-E dataset and outperformed all type-I and type-II approaches.
Compared to the poor performance of the \emph{vanilla} molecular docking method, 
all the deep learning-based approaches achieved much higher AUROC.
To eliminate the effect of data splitting schemes, 
we further selected the state-of-the-art type-II approach
MONN~\citep{li2020monn} and the state-of-the-art type-I approach \citet{lim2019gnndti}
(referred to as GNN-DTI in the following text for convenience) for
reproduction with the same data splitting scheme during model training and evaluation.
We also evaluated the performance on
several metrics besides AUROC, i.e., AUPRC, adjusted LogAUC, ROC enrichment, and enrichment factor,
to avoid the potential bias of a single metric.
As shown in Table~\ref{tbl:dude-litpcba-perf}, IGT achieved the best performance in terms of all the six metrics, 
which shows its superior capability for DTI prediction.

\begin{table*}
  \caption{
    Performance evaluation of binding activity prediction on DUD-E and LIT-PCBA.
  }
  \label{tbl:dude-litpcba-perf}
  \centering
  \begin{tabular}{lccccccc}
    \hline
    Dataset & Model                                                    & AUROC & LogAUC & AUPRC & Bal. Acc. & ROC Enrich. & Enrich. Factor \\
    \hline
    \multirow{4}{*}{DUD-E}
    & IGT                                            & \textbf{0.981} & \textbf{0.754} & \textbf{0.730} & \textbf{0.875} & \textbf{75.9} & \textbf{36.6} \\
    & MONN                               & 0.969 & 0.705 & 0.669 & 0.768 & 66.2 & 32.8 \\
    & GNN-DTI                          & 0.960 & 0.668 & 0.581 & 0.864 & 58.7 & 29.7 \\
    & Docking                                                  & 0.707 & 0.156 & 0.139 &   -   & 10.4 & 6.35\\
    \hline
    \multirow{4}{*}{LIT-PCBA}
    & IGT             & \textbf{0.942} & \textbf{0.586} & \textbf{0.311} & \textbf{0.871} & \textbf{40.1} & 23.3 \\
    & MONN            & 0.940 & 0.582 & 0.286 & 0.861 & 38.6 & \textbf{23.4} \\
    & GNN-DTI        & 0.925 & 0.549 & 0.253 & 0.839 & 32.2 & 20.4 \\
    & Docking         & 0.558 & 0.035 & 0.013 &   -   & 2.36 & 1.83 \\
    \hline
  \end{tabular}
\end{table*}

Although DUD-E is widely adopted as a benchmark dataset, 
it is reported to be biased~\citep{xia2015benchmarking,chen2019hidden,tran2020lit},
which likely leads to poor generalization capability of the models.
\citet{xia2015benchmarking} illustrated three types of biases,
i.e., analogue bias, artificial enrichment, and false negatives. These
biases are commonly existed in the virtual screening datasets and thus
lead to over-optimistic estimates of model performance and poor
generalization of the models.  For
instance, \citet{chen2019hidden} pointed out that, on the DUD-E
dataset, the models trained without any information of target proteins
can still achieve comparable performance of the state-of-the-art
algorithms, showing that the models overfitted to dataset biases
instead of learning the physical rules for the
predictions.  To alleviate the biases in the
datasets, \citet{tran2020lit} extracted experimentally validated
protein-ligand binding pairs from PubChem~\citep{kim2021pubchem} and
filtered the data with stringent criteria.  The
resultant dataset, LIT-PCBA, was reported to be unbiased, which is
favored for training deep learning models for DTI prediction.

Therefore, we employed the LIT-PCBA dataset for model training and evaluation. 
Following the same procedures for DUD-E, we also split LIT-PCBA into training, validation, and test sets, 
and trained IGT with the identical hyper-parameters. 
In addition, MONN and GNN-DTI were also trained on the same dataset with their default hyper-parameters.
As shown in Table~\ref{tbl:dude-litpcba-perf}, we compared the performance of IGT with that of MONN and GNN-DTI.
The \emph{vanilla} molecular docking method only achieved a 0.558 AUROC score on LIT-PCBA, 
significantly lower than that on DUD-E and the other three approaches on LIT-PCBA. 
Our IGT model performed the best in terms of five metrics out of the six. %

\subsection{IGT generalizes better to unseen receptors and compounds}

\begin{table*}[ht]
  \caption{
    Evaluation of the generalization ability among IGT, MOON and GNN-DTI on MUV test set.
  }
  \label{tbl:muv-perf}
  \centering
  \begin{tabular}{lcc}
    \hline
    Model & AUROC (trained on DUD-E)  & AUROC (trained on LIT-PCBA) \\
    \hline
    IGT & \textbf{0.547} &  \textbf{0.634} \\
    MONN &  0.546 & 0.560 \\
    GNN-DTI &  0.540 & 0.602 \\
    \hline
  \end{tabular}
\end{table*}

IGT has achieved superior performance on both DUD-E and LIT-PCBA.
However, those results do not necessarily
reflect the performance of the models on real DTI applications since unseen receptors and/or compounds are often
involved in the real-world scenarios, such as drug virtual screening for novel targets. 
Therefore, we further evaluated the models against MUV, a new
test dataset that consists of different receptor proteins from DUD-E and LIT-PCBA.
As demonstrated above, we trained IGT and reproduced MOON and GNN-DTI with their default hyper-parameters on DUD-E and LIT-PCBA, respectively, 
resulting in a total of six models for binding activity prediction.
All these models were evaluated on the MUV test set, and their performance is compared in Table~\ref{tbl:muv-perf} and Fig. S1.

Two observations can be drawn from these results.
First, for all these three approaches, the performance of a prediction model trained on LIT-PCBA dataset 
was better than that of the same model trained on DUD-E dataset, 
indicating that training on unbiased samples does increase the model performance on
unseen receptors and compounds.
Second, although performance gains were observed in all the three approaches when trained on LIT-PCBA, 
the improvement for MONN is only 0.014, which is relatively small compared with 0.087 for IGT and 0.062 for GNN-DTI.
This implies that even if a DTI model is trained on unbiased data, its generalization capability still depends a lot on model design.

This result shows that IGT is much more ready for real-world DTI applications because of its superior
generalization ability.  However, one might question about what leads to IGT's better performance.
To answer this question, we conduct ablation studies as reported in the next section.

\subsection{Ablation study}

\begin{figure*}%
  \centering
  \includegraphics[width=.9\textwidth]{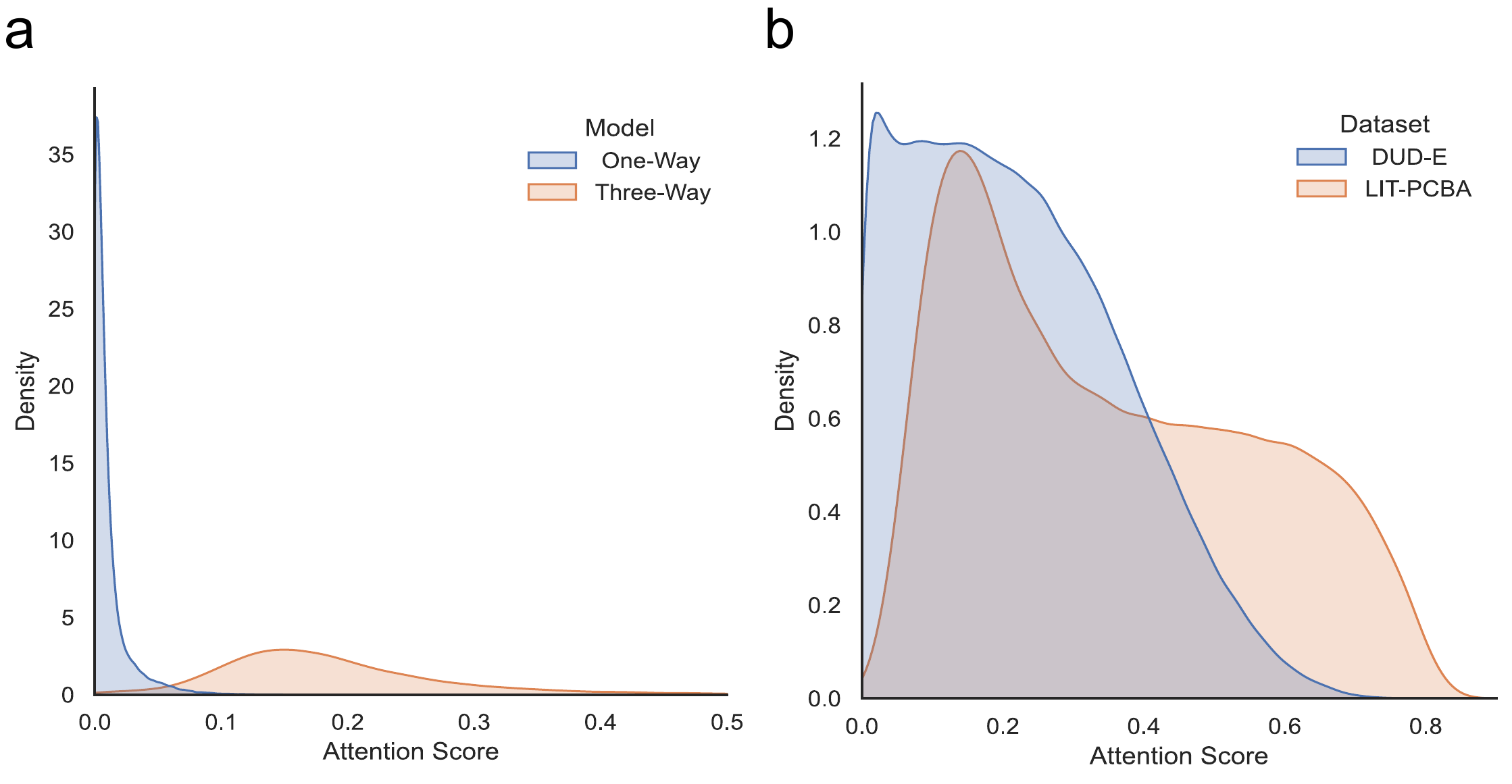}
  \caption{
    The distributions of mean intermolecular attention scores.
    \textbf{a}. Comparison between the distribution of the three-way model and that of the one-way model.  The two models were both trained and evaluated on LIT-PCBA training and test sets.
    \textbf{b}. Comparison between the distribution of the three-way models on MUV test set.  The two models were trained on DUD-E and LIT-PCBA datasets, respectively.
  }
  \label{fig:attention-score}
\end{figure*}

To better understand where the performance gains come from, 
we conducted an extensive ablation study 
on both the features used in IGT and the model architecture we designed.
Different sets of features or model components were removed from the models, and the resulting models were train
on the LIT-PCBA dataset and evaluated on the validation set with AUROC.

\noindent \textbf{Features.}
As shown in Table S1, we first removed all node features except the basic atom symbol, 
resulted in a 0.030 decrease of the AUROC score. 
This result indicates that the atom features do contribute to the model performance significantly as expected.
We then tried to remove the feature of intermolecular distances, which also led to a large performance drop. 
In contrast, removing all edge features but the intermolecular distances resulted in only a small decrease of AUROC score.
This indicates that the inter-atomic distances are probably much more important than the other edge features.
Furthermore, performance drops were observed in all the three experiments, 
showing that both atom and edge features contribute to the DTI predictions.
The ablation study on features shows that the IGT can learn from both the docked structure (e.g., the distance feature)
and the physicochemical properties of the individual atoms.

\noindent \textbf{Modules.}
We further conducted the ablation study on the neural network architecture of the model.
We devised two experiments by modifying the model architecture to study how the model learns intermolecular information.
When the intermolecular attentions were removed (i.e., resulting in a one-way graph transformer), a 0.016 performance drop was observed.
For the complex graph in each building block, removing intramolecular edge features resulted in a similar performance drop.
This result demonstrates that both the modelling of the separated components and that of their interactions are important to the DTI prediction.

\noindent \textbf{Three-way versus one-way.}
We further investigate the effect of the intermolecular attention and the three-way design.
To figure out whether such design can truly improve the model's ability to capture intermolecular
interactions, we compare the IGT with its one-way counterpart, using the ``unsupervised dot-product attention scores''.
The unsupervised dot-product attention scores are calculated as in Eq.~\ref{eq:attention-score}.

\begin{equation}
  A=\texttt{softmax}(HH^T) \label{eq:attention-score}
\end{equation}
where $H\in \mathbf{R}^{N\times d}$ denotes the graph node embedding tensor.
The entry $A_{ij}$ can be seen as the contribution of node $j$ to node $i$.
We calculate the attention scores $A$ right before the final prediction layers for both the three-way
and one-way models, and then compute the mean of the entries of $A$ that correspond to intermolecular
interactions (i.e., $A_{ij}$ such that $i$ belongs to ligand and $j$ belongs to receptor, and vice versa).
This score, which we call the ``mean intermolecular attention score'', reflects how much the intermolecular interactions contribute to the node representations.
The procedure is repeated for all proteins in the test set to obtain a distribution.
As shown in Fig.~\ref{fig:attention-score}a, the distributions for the two models are significantly
different.  Concretely, the distribution for the three-way model is much more wider and right-shifted than that of the
one-way model, indicating that the prediction in the three-way model relies more on intermolecular information.
This confirms that the three-way design, as well as the intermolecular attention mechanism,
greatly improves the modeling of intermolecular interactions, explaining the better
performance of IGT.

\noindent \textbf{DUD-E versus LIT-PCBA.}
As mentioned in previous sections, it is well acknowledged that the DUD-E dataset contains various
types of biases and is not suitable to serve as a training set for machine learning.
In this section, we quantitatively analyze this problem by comparing the distributions of mean intermolecular attention
scores for two IGT models, one trained on DUD-E and the other trained on LIT-PCBA.
As shown in Fig.~\ref{fig:attention-score}b, the distribution for the LIT-PCBA model is much more right-shifted
than the one of the DUD-E model, meaning that the model trained on LIT-PCBA pays much more attention
to the intermolecular interactions.
This provides an explanation of IGT's superb generalization ability over previous models:
the correct choice of an unbiased training dataset and the reasonable model design of IGT jointly allow for
the successful learning of intermolecular information, which is the key to generalization.

\subsection{IGT successfully identifies best poses in pose prediction}

Searching the binding pose for a given complex closest to the native conformation is essential to 
understanding intermolecular interactions. %
We evaluated the performance of IGT for binding pose prediction based on a refined set of the PDBbind database~\citep{liu2017pdbbind}.
We used Smina to generate multiple conformations of the same complex with \emph{exhaustiveness}=50 and \emph{num modes}=20.
We then calculated the \emph{root-mean-square deviation} (RMSD) for the candidates against its crystal structure.
In this way, candidate poses are labeled either positive or negative according to their RMSDs with the crystal structure.
Specifically, a positive pose has an RMSD less than 2 \angstrom{} and a negative pose has an RMSD larger than 4 \angstrom.
We then randomly split the dataset into training, validation, and test sets with a ratio of 0.70:0.15:0.15.
We used the identical model architecture as the one we used for binding activity prediction and trained IGT on the training set with a learning rate of 1e-5.
We then evaluated the AUROC and the PRAUC scores of IGT on the test set. 
MONN is a type-II model and cannot be applied to this task, since all candidate poses share the same input representations.
As shown in Table~\ref{tbl:pdbdind-perf}, the performance of both type-I models 
for binding pose prediction is much better than that of the vanilla molecular docking method.
Compared with the best type-I approach GNN-DTI~\citep{lim2019gnndti}, 
the AUROC and PRAUC scores of our model are 0.024 and 0.046 higher, respectively.
As mentioned in binding activity prediction, 
our IGT model is capable of modelling the structures of the complexes and therefore improves the performance on binding pose prediction.

Fig. S2 shows the percentages of poses whose RMSD are smaller than 2\angstrom 
within the top-K poses ranked by IGT, GNN-DTI, and the docking method, respectively.
Compared with the docking method and GNN-DTI, our model shows greater ability to identify near-native poses.

\subsection{Applying IGT to SARS-CoV-2}

\begin{figure}%
  \centering
  \includegraphics[width=0.44\textwidth]{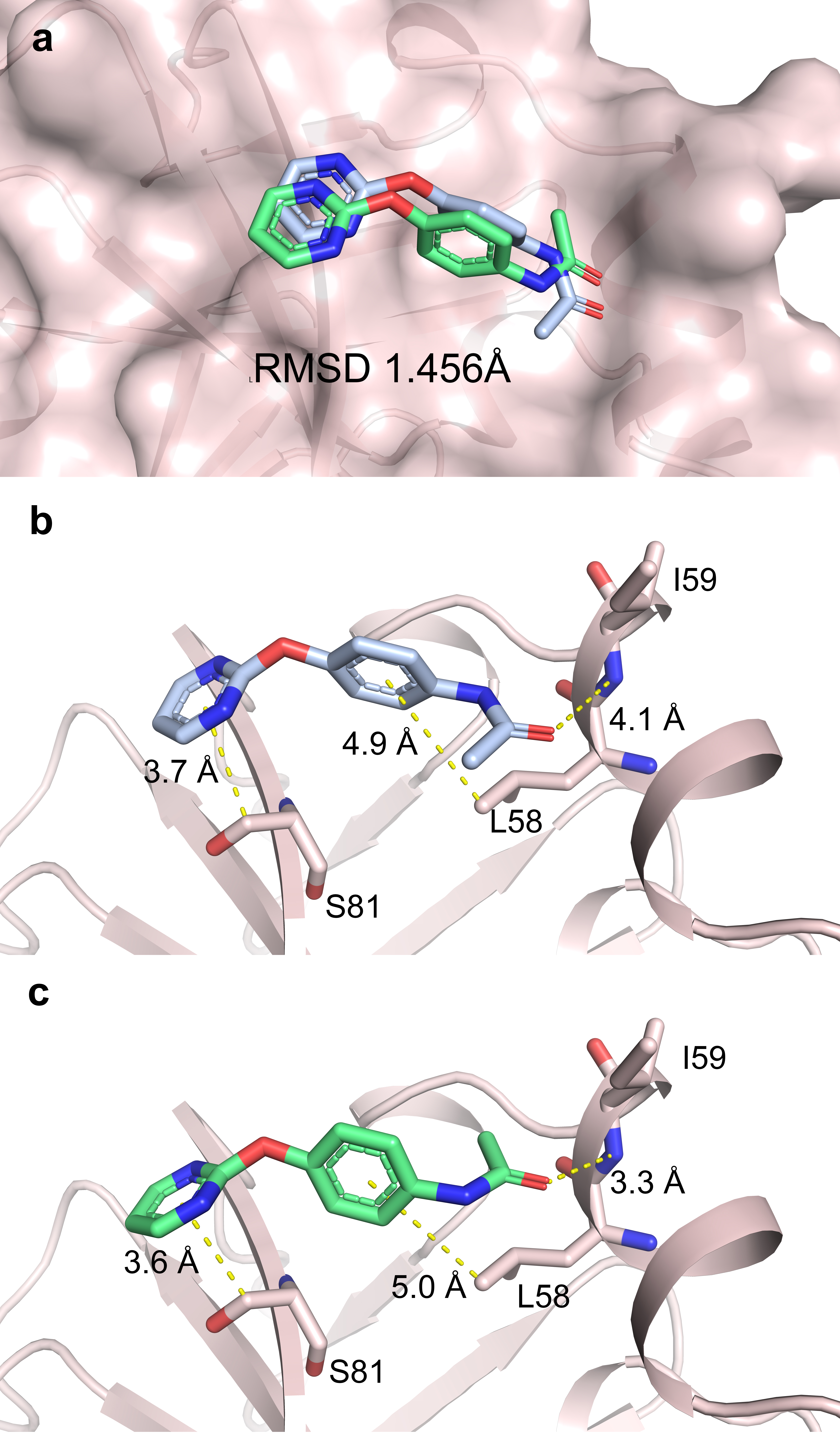}
  \caption{
    Case study for drug virtual screening against SARS-CoV-2 by IGT model.
    \textbf{a}. The docked poses of the drug-protein complex. The predicted binding pose and the ground truth are colored green and lilac, respectively. The binding pocket of the receptor protein is colored amaranth with surface. The RMSD value between the predicted binding pose and the ground truth is 1.456 \angstrom.
    \textbf{b}. Interaction analysis between the ground truth ligand pose and the protein.
    \textbf{c}. Interaction analysis between the predicted binding pose and the receptor.
    In b and c, the ligands as well as interactive residues are shown as sticks and the remaining protein structures are shown as secondary structures.
  }
  \label{fig:sars-cov-2}
\end{figure}

To further validate the practical value of IGT, the Diamond SARS-CoV-2 drug dataset was used as a real-world evaluation.
We prepared two models of IGT for this task, i.e., the binding activity prediction model (M1) and the binding pose prediction model (M2).
The M1 model was trained on LIT-PCBA while M2 was trained on PDBbind.
After molecular docking, Smina was run to generate binding poses.
M1 was then used to predict the activity of ligands against the SARS-CoV-2 main protease.
As a result, the IGT M1 model successfully recalled 83.12$\%$ active ligands and achieved the AUROC score of 0.701.
Then we selected the active binding ligands that were also predicted as the active ones by M1 model and identified the best binding pose for each such protein-ligand pair using the M2 model. 
As shown in Figure~\ref{fig:sars-cov-2}, the predicted binding pose selected by the M2 model is basically indistinguishable from the native structure with a RMSD of 1.456 \angstrom, indicating that our IGT model successfully detects the near-native pose against SARS-CoV-2. 
Furthermore, the predicted binding pose formed similar interactions that can also be found in the native complex structure. 
In particular, in the predicted drug binding pose, a $\pi$-$\sigma$ interaction and a hydrophobic interaction were formed between S81 and L58, respectively.
Furthermore, a stronger hydrogen bond was formed with the residue I59 compared with that in the native pose, indicating a more stable binding to the pocket of the receptor protein.

\begin{table}
  \caption{
    Performance evaluation of binding pose prediction on PDBbind.
  }
  \label{tbl:pdbdind-perf}
  \centering
  \begin{tabular}{lcc}
    \hline
    Model                                                    & AUROC & AUPRC \\
    \hline
    IGT                                                & \textbf{0.915} & \textbf{0.765} \\
    GNN-DTI                           & 0.854 & 0.635 \\
    Docking                                                  & 0.702 & 0.466 \\
    \hline
  \end{tabular}
\end{table}

The evaluation on SARS-CoV-2 drug dataset shows that our model not only performs well on the benchmark datasets like DUD-E, LIT-PCBA or MUV,
but also has its practical value to be applied to real-world DTI scenarios.

\section{Discussion and Conclusion}

We have demonstrated the superior performance of IGT in terms of both fitting and generalization capabilities
through the evaluations on the DUD-E, LIT-PCBA, MUV, and the SARS-CoV-2 main protease DTI datasets.
The factors that contribute to the superior performance of IGT can be summarized as follows.
First, the architecture of IGT assigns more weights to intermolecular information than intramolecular information,
so it can better reflect the physical rules in drug-target interactions as well as avoid fitting to the biases in the data.
Second, the design of IGT is more robust to the spatial relationship between the ligand and the receptor.
For example, the predictions of IGT are invariant to translations and rotations of the inputs (\textit{translational and rotational equivariance})
because it only leverages features invariant to such transformations. 
In contrast, this deserved property is seldom found in the voxelization-based methods (e.g., 3D CNNs).
Third, training on the unbiased LIT-PCBA dataset allows IGT to fit to the signal rather than the biases in the data.

Although IGT performs well on both binding activity prediction and pose prediction, it still
has a few limitations.  First, IGT requires the docking poses from molecular docking software
as input. This may lead to error propagation since the molecular docking procedure itself is non-deterministic
and inaccurate.  Second, molecular docking slows down the execution of IGT, hindering its efficiency of virtually screening active drugs from millions to billions chemical compounds.
Third, although IGT exhibits superior generalization ability to unseen receptors and compounds, 
the performance is still much lower than that evaluated on similar test sets. 
Designing a DTI prediction model with better generalization ability and lower resource consumption, still awaits future study.

\balance

\bibliographystyle{unsrtnat}
\bibliography{paper}%

\begin{thebibliography}{29}
\providecommand{\natexlab}[1]{#1}
\providecommand{\url}[1]{\texttt{#1}}
\expandafter\ifx\csname urlstyle\endcsname\relax
  \providecommand{\doi}[1]{doi: #1}\else
  \providecommand{\doi}{doi: \begingroup \urlstyle{rm}\Url}\fi

\bibitem[da~Silva~Rocha et~al.(2019)da~Silva~Rocha, Olanda, Fokoue, and
  Sant'Anna]{da2019virtual}
Sheisi~FL da~Silva~Rocha, Carolina~G Olanda, Harold~H Fokoue, and Carlos~MR
  Sant'Anna.
\newblock Virtual screening techniques in drug discovery: review and recent
  applications.
\newblock \emph{Current topics in medicinal chemistry}, 19\penalty0
  (19):\penalty0 1751--1767, 2019.

\bibitem[Lin et~al.(2020)Lin, Li, and Lin]{lin2020review}
Xiaoqian Lin, Xiu Li, and Xubo Lin.
\newblock A review on applications of computational methods in drug screening
  and design.
\newblock \emph{Molecules}, 25\penalty0 (6):\penalty0 1375, 2020.

\bibitem[Wu et~al.(2020)Wu, Liu, Yang, Zhang, Zhong, Wang, Wang, Xu, Li, Li,
  et~al.]{wu2020analysis}
Canrong Wu, Yang Liu, Yueying Yang, Peng Zhang, Wu~Zhong, Yali Wang, Qiqi Wang,
  Yang Xu, Mingxue Li, Xingzhou Li, et~al.
\newblock Analysis of therapeutic targets for sars-cov-2 and discovery of
  potential drugs by computational methods.
\newblock \emph{Acta Pharmaceutica Sinica B}, 10\penalty0 (5):\penalty0
  766--788, 2020.

\bibitem[Ashburn and Thor(2004)]{ashburn2004drug}
Ted~T Ashburn and Karl~B Thor.
\newblock Drug repositioning: identifying and developing new uses for existing
  drugs.
\newblock \emph{Nature reviews Drug discovery}, 3\penalty0 (8):\penalty0
  673--683, 2004.

\bibitem[Novac(2013)]{novac2013challenges}
Natalia Novac.
\newblock Challenges and opportunities of drug repositioning.
\newblock \emph{Trends in pharmacological sciences}, 34\penalty0 (5):\penalty0
  267--272, 2013.

\bibitem[Xue et~al.(2018)Xue, Li, Xie, and Wang]{xue2018review}
Hanqing Xue, Jie Li, Haozhe Xie, and Yadong Wang.
\newblock Review of drug repositioning approaches and resources.
\newblock \emph{International journal of biological sciences}, 14\penalty0
  (10):\penalty0 1232, 2018.

\bibitem[Dudley et~al.(2011)Dudley, Deshpande, and Butte]{dudley2011exploiting}
Joel~T Dudley, Tarangini Deshpande, and Atul~J Butte.
\newblock Exploiting drug--disease relationships for computational drug
  repositioning.
\newblock \emph{Briefings in bioinformatics}, 12\penalty0 (4):\penalty0
  303--311, 2011.

\bibitem[Csermely et~al.(2013)Csermely, Korcsm{\'a}ros, Kiss, London, and
  Nussinov]{csermely2013structure}
Peter Csermely, Tam{\'a}s Korcsm{\'a}ros, Huba~JM Kiss, G{\'a}bor London, and
  Ruth Nussinov.
\newblock Structure and dynamics of molecular networks: a novel paradigm of
  drug discovery: a comprehensive review.
\newblock \emph{Pharmacology \& therapeutics}, 138\penalty0 (3):\penalty0
  333--408, 2013.

\bibitem[De~Vivo et~al.(2016)De~Vivo, Masetti, Bottegoni, and
  Cavalli]{de2016role}
Marco De~Vivo, Matteo Masetti, Giovanni Bottegoni, and Andrea Cavalli.
\newblock Role of molecular dynamics and related methods in drug discovery.
\newblock \emph{Journal of medicinal chemistry}, 59\penalty0 (9):\penalty0
  4035--4061, 2016.

\bibitem[Durrant and McCammon(2011)]{durrant2011molecular}
Jacob~D Durrant and J~Andrew McCammon.
\newblock Molecular dynamics simulations and drug discovery.
\newblock \emph{BMC biology}, 9\penalty0 (1):\penalty0 1--9, 2011.

\bibitem[{\'S}led{\'z} and Caflisch(2018)]{sledz2018protein}
Pawe{\l} {\'S}led{\'z} and Amedeo Caflisch.
\newblock Protein structure-based drug design: from docking to molecular
  dynamics.
\newblock \emph{Current opinion in structural biology}, 48:\penalty0 93--102,
  2018.

\bibitem[Lim et~al.(2019)Lim, Ryu, Park, Choe, Ham, and Kim]{lim2019gnndti}
Jaechang Lim, Seongok Ryu, Kyubyong Park, Yo~Joong Choe, Jiyeon Ham, and
  Woo~Youn Kim.
\newblock Predicting drug--target interaction using a novel graph neural
  network with 3d structure-embedded graph representation.
\newblock \emph{Journal of chemical information and modeling}, 59\penalty0
  (9):\penalty0 3981--3988, 2019.

\bibitem[Ragoza et~al.(2017)Ragoza, Hochuli, Idrobo, Sunseri, and
  Koes]{ragoza2017protein}
Matthew Ragoza, Joshua Hochuli, Elisa Idrobo, Jocelyn Sunseri, and David~Ryan
  Koes.
\newblock Protein--ligand scoring with convolutional neural networks.
\newblock \emph{Journal of chemical information and modeling}, 57\penalty0
  (4):\penalty0 942--957, 2017.

\bibitem[Wallach et~al.(2015)Wallach, Dzamba, and Heifets]{wallach2015atomnet}
Izhar Wallach, Michael Dzamba, and Abraham Heifets.
\newblock Atomnet: a deep convolutional neural network for bioactivity
  prediction in structure-based drug discovery.
\newblock \emph{arXiv preprint arXiv:1510.02855}, 2015.

\bibitem[Mysinger et~al.(2012)Mysinger, Carchia, Irwin, and
  Shoichet]{mysinger2012dude}
Michael~M Mysinger, Michael Carchia, John~J Irwin, and Brian~K Shoichet.
\newblock Directory of useful decoys, enhanced (dud-e): better ligands and
  decoys for better benchmarking.
\newblock \emph{Journal of medicinal chemistry}, 55\penalty0 (14):\penalty0
  6582--6594, 2012.

\bibitem[Li et~al.(2020)Li, Wan, Shu, Jiang, Zhao, and Zeng]{li2020monn}
Shuya Li, Fangping Wan, Hantao Shu, Tao Jiang, Dan Zhao, and Jianyang Zeng.
\newblock Monn: a multi-objective neural network for predicting
  compound-protein interactions and affinities.
\newblock \emph{Cell Systems}, 10\penalty0 (4):\penalty0 308--322, 2020.

\bibitem[Zheng et~al.(2020)Zheng, Li, Chen, Xu, and Yang]{zheng2020drugvqa}
Shuangjia Zheng, Yongjian Li, Sheng Chen, Jun Xu, and Yuedong Yang.
\newblock Predicting drug--protein interaction using quasi-visual
  questionanswering system.
\newblock \emph{Nature Machine Intelligence}, 2\penalty0 (2):\penalty0
  134--140, Feb 2020.
\newblock ISSN 2522-5839.
\newblock \doi{10.1038/s42256-020-0152-y}.
\newblock URL \url{https://doi.org/10.1038/s42256-020-0152-y}.

\bibitem[Torng and Altman(2019)]{torng2019graph}
Wen Torng and Russ~B Altman.
\newblock Graph convolutional neural networks for predicting drug-target
  interactions.
\newblock \emph{Journal of chemical information and modeling}, 59\penalty0
  (10):\penalty0 4131--4149, 2019.

\bibitem[Vaswani et~al.(2017)Vaswani, Shazeer, Parmar, Uszkoreit, Jones, Gomez,
  Kaiser, and Polosukhin]{vaswani2017attention}
Ashish Vaswani, Noam Shazeer, Niki Parmar, Jakob Uszkoreit, Llion Jones,
  Aidan~N Gomez, Lukasz Kaiser, and Illia Polosukhin.
\newblock Attention is all you need.
\newblock \emph{arXiv preprint arXiv:1706.03762}, 2017.

\bibitem[Dwivedi and Bresson(2021)]{dwivedi2021generalization}
Vijay~Prakash Dwivedi and Xavier Bresson.
\newblock A generalization of transformer networks to graphs, 2021.

\bibitem[Tran-Nguyen et~al.(2020)Tran-Nguyen, Jacquemard, and
  Rognan]{tran2020lit}
Viet-Khoa Tran-Nguyen, C{\'e}lien Jacquemard, and Didier Rognan.
\newblock Lit-pcba: An unbiased data set for machine learning and virtual
  screening.
\newblock \emph{Journal of chemical information and modeling}, 60\penalty0
  (9):\penalty0 4263--4273, 2020.

\bibitem[Rohrer and Baumann(2009)]{rohrer2009maximum}
Sebastian~G. Rohrer and Knut Baumann.
\newblock Maximum unbiased validation (muv) data sets for virtual screening
  based on pubchem bioactivity data.
\newblock \emph{Journal of Chemical Information and Modeling}, 49\penalty0
  (2):\penalty0 169--184, 2009.
\newblock \doi{10.1021/ci8002649}.
\newblock URL \url{https://doi.org/10.1021/ci8002649}.
\newblock PMID: 19161251.

\bibitem[Liu et~al.(2017)Liu, Su, Han, Liu, Yang, Li, and Wang]{liu2017pdbbind}
Zhihai Liu, Minyi Su, Li~Han, Jie Liu, Qifan Yang, Yan Li, and Renxiao Wang.
\newblock Forging the basis for developing protein--ligand interaction scoring
  functions.
\newblock \emph{Accounts of chemical research}, 50\penalty0 (2):\penalty0
  302--309, 2017.

\bibitem[Dosovitskiy et~al.(2020)Dosovitskiy, Beyer, Kolesnikov, Weissenborn,
  Zhai, Unterthiner, Dehghani, Minderer, Heigold, Gelly,
  et~al.]{dosovitskiy2020image}
Alexey Dosovitskiy, Lucas Beyer, Alexander Kolesnikov, Dirk Weissenborn,
  Xiaohua Zhai, Thomas Unterthiner, Mostafa Dehghani, Matthias Minderer, Georg
  Heigold, Sylvain Gelly, et~al.
\newblock An image is worth 16x16 words: Transformers for image recognition at
  scale.
\newblock \emph{arXiv preprint arXiv:2010.11929}, 2020.

\bibitem[Koes et~al.(2013)Koes, Baumgartner, and Camacho]{koes2013lessons}
D.~R. Koes, M.~P. Baumgartner, and C.~J. Camacho.
\newblock Lessons learned in empirical scoring with smina from the csar 2011
  benchmarking exercise.
\newblock \emph{J Chem Inf Model}, 53\penalty0 (8):\penalty0 1893--904, 2013.
\newblock ISSN 1549-960X (Electronic) 1549-9596 (Linking).
\newblock \doi{10.1021/ci300604z}.
\newblock URL \url{https://www.ncbi.nlm.nih.gov/pubmed/23379370}.

\bibitem[Trott and Olson(2010)]{trott2010vina}
O.~Trott and A.~J. Olson.
\newblock Autodock vina: improving the speed and accuracy of docking with a new
  scoring function, efficient optimization, and multithreading.
\newblock \emph{J Comput Chem}, 31\penalty0 (2):\penalty0 455--61, 2010.
\newblock ISSN 1096-987X (Electronic) 0192-8651 (Linking).
\newblock \doi{10.1002/jcc.21334}.
\newblock URL \url{https://www.ncbi.nlm.nih.gov/pubmed/19499576}.

\bibitem[Xia et~al.(2015)Xia, Tilahun, Reid, Zhang, and
  Wang]{xia2015benchmarking}
Jie Xia, Ermias~Lemma Tilahun, Terry-Elinor Reid, Liangren Zhang, and
  Xiang~Simon Wang.
\newblock Benchmarking methods and data sets for ligand enrichment assessment
  in virtual screening.
\newblock \emph{Methods}, 71:\penalty0 146--157, 2015.
\newblock \doi{https://doi.org/10.1016/j.ymeth.2014.11.015}.
\newblock URL
  \url{https://www.sciencedirect.com/science/article/pii/S1046202314003788}.

\bibitem[Chen et~al.(2019)Chen, Cruz, Ramsey, Dickson, Duca, Hornak, Koes, and
  Kurtzman]{chen2019hidden}
Lieyang Chen, Anthony Cruz, Steven Ramsey, Callum~J Dickson, Jose~S Duca,
  Viktor Hornak, David~R Koes, and Tom Kurtzman.
\newblock Hidden bias in the dud-e dataset leads to misleading performance of
  deep learning in structure-based virtual screening.
\newblock \emph{PloS one}, 14\penalty0 (8):\penalty0 e0220113, 2019.

\bibitem[Kim et~al.(2020)Kim, Chen, Cheng, Gindulyte, He, He, Li, Shoemaker,
  Thiessen, Yu, Zaslavsky, Zhang, and Bolton]{kim2021pubchem}
Sunghwan Kim, Jie Chen, Tiejun Cheng, Asta Gindulyte, Jia He, Siqian He,
  Qingliang Li, Benjamin~A Shoemaker, Paul~A Thiessen, Bo~Yu, Leonid Zaslavsky,
  Jian Zhang, and Evan~E Bolton.
\newblock {PubChem in 2021: new data content and improved web interfaces}.
\newblock \emph{Nucleic Acids Research}, 49\penalty0 (D1):\penalty0
  D1388--D1395, 11 2020.
\newblock ISSN 0305-1048.
\newblock \doi{10.1093/nar/gkaa971}.
\newblock URL \url{https://doi.org/10.1093/nar/gkaa971}.

\end{thebibliography}

\hfill %
\end{document}